\documentclass[sigconf]{acmart}

\usepackage{booktabs} 
\usepackage{multirow}
\usepackage{algorithm, url}
\usepackage[noend]{algorithmic}

\copyrightyear{2017}
\acmYear{2017}
\setcopyright{acmcopyright}
\acmConference{SIGIR '17}{August 07-11, 2017}{Shinjuku, Tokyo, Japan}
\acmPrice{15.00}
\acmDOI{10.1145/3077136.3080655} 
\acmISBN{978-1-4503-5022-8/17/08}

\begin{document}
\title{Video Question Answering via Attribute-Augmented Attention Network Learning}

\author{Yunan Ye \; Zhou Zhao \; Yimeng Li \; Long Chen \; Jun Xiao}
\authornote{Corresponding author.}
\author{Yueting Zhuang}
\affiliation{%
  \institution{College of Computer Science, Zhejiang University, China}
}
\email{{chryleo, zhaozhou, aquaird, longc, junx, yzhuang}@zju.edu.cn}

\begin{abstract}
Video Question Answering is a challenging problem in visual information retrieval, which provides the answer to the referenced video content according to the question. However, the existing visual question answering approaches mainly tackle the problem of static image question, which may be ineffectively for video question answering due to the insufficiency of modeling the temporal dynamics of video contents. In this paper, we study the problem of video question answering by modeling its temporal dynamics with frame-level attention mechanism. We propose the attribute-augmented attention network learning framework that enables the joint frame-level attribute detection and unified video representation learning for video question answering. We then incorporate the multi-step reasoning process for our proposed attention network to further improve the performance. We construct a large-scale video question answering dataset. We conduct the experiments on both multiple-choice and open-ended video question answering tasks to show the effectiveness of the proposed method.
\end{abstract}



\maketitle

\section{Introduction}
Visual information retrieval (VIR) is the information delivery mechanism that enables users to post their queries and then obtain the answers from visual contents~\cite{gupta1997visual}. As an emerging kind of recommender system, visual question answering is an important problem for VIR sites, which automatically returns the relevant answer from the referenced visual contents according to users' posted question~\cite{antol2015vqa, HeZKC16, HeGKW17, HeLZNHC17}. Currently, most of the existing visual question answering methods mainly focus on the problem of static image question answering~\cite{antol2015vqa, yang2016stacked, NieWGZC13, LuoNYW16, NieWZLC11, WangLTLW12}. Although existing methods have achieved promising performance in image question answering task, they may still be ineffective applied to the problem of video question answering due to the lack of modeling the temporal dynamics of video contents~\cite{ZhaoYCHZ16, ZhaoLZCHZ17}.

\begin{figure}
\centering
\includegraphics[width=0.35\textwidth,height=0.12\textwidth]{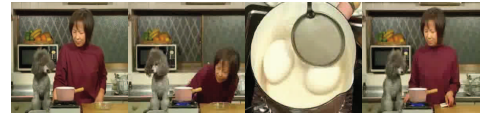}
\caption{Video Question Answering}\label{fig:demo}
\end{figure}

The video content often contains the evolving complex interactions and the simple extension of image question answering is thus ineffectively to provide the satisfactory answers. This is because the relevant video information is usually scattered among the entire frames. Furthermore, a number of frames in video are redundant and irrelevant to the question. We give a simple example of video question answering in Figure~\ref{fig:demo}. We demonstrate that the answering for question "What is a woman boiling in a pot of water?" requires the collective information from multiple video frames. Recently, temporal attention mechanisms have been shown to its effectiveness on critical frame extraction for video representation learning~\cite{WangHLZYC12}.
Thus, we then employ the temporal attention mechanisms to model the temporal dynamics of video contents. On the other hand, the utilization of high-level semantic attributes has demonstrated the effectiveness in visual understanding tasks~\cite{NieYWHC12}. Furthermore, we observe that the detected attributes are able to enhance the performance of video question answering in Figure~\ref{fig:demo}. Thus, leveraging both temporal dynamic modeling and semantic attributes is critical for learning effective video representation in video question answering.

\begin{figure*}[t]
\centering
\includegraphics[width=0.7\textwidth]{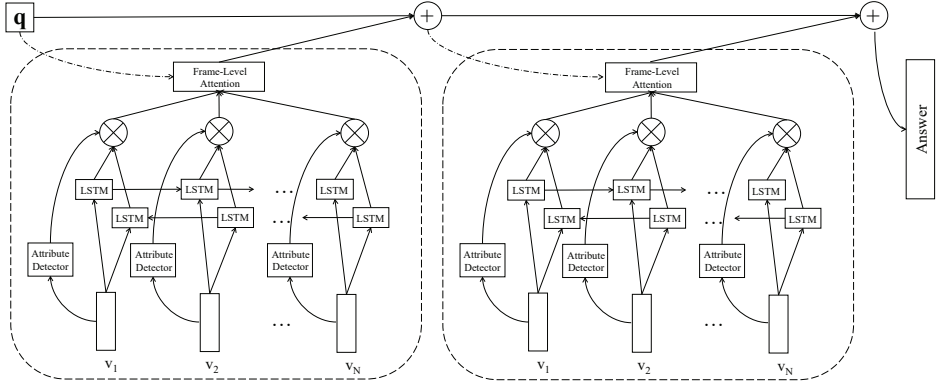}
\caption{The Overview of Video Question Answering via Attribute-Augmented Attention Network Learning}\label{fig:framework}
\end{figure*}
In this paper, we study the problem of video question answering by modeling its temporal dynamics and semantic attributes. Specifically, we propose the attribute-augmented attention network learning framework that enables the joint frame-level attribute detection and unified video representation learning for video question answering. We then incorporate the multi-step reasoning process for our proposed attribute-augmented attention network to further improve the performance, named as r-ANL. When a certain question is issued, r-ANL can return the relevant answer for it based on the referenced video content. The main contributions of this paper are as follows:
\begin{itemize}
\item Unlike the previous studies, we study the problem of video question answering by modeling its temporal dynamics and semantic attributes. We propose the attribute-augmented attention network learning framework that jointly detects frame-level attribute and learns the unified video representation for video question answering.
\item We incorporate the multi-step reasoning process for the proposed attention networks to enable the progressive joint representation learning of multimodal temporal attentional video with semantic attributes and textual question to further improve the performance of video question answering.
\item We construct a large-scale dataset for video question answering. We evaluate the performance of our method on both multiple choice and open-ended video question answering tasks.
\end{itemize}

\section{Video Question Answering via Attention Network Learning}

\subsection{Problem Formulation}
Before presenting our method, we first introduce some basic notions and terminologies. We denote the question by ${\bf q}\in Q$, the video by ${\bf v}\in V$ and the attributes by ${\bf a}\in A$, respectively. The frame-level representation of video ${\bf v}$ is given by ${\bf v}=({\bf v}_{1},{\bf v}_{2},\ldots,{\bf v}_{N})$, where $N$ is the length of video ${\bf v}$. We then denote the frame-level representation of attribute ${\bf a}_{v}$ for video ${\bf v}$ by ${\bf a}_{v}=(a_{v,1},a_{v,2},\ldots,a_{v,N})$, where $a_{v,i}$ is the set of the attributes for the $i$-th frame. We then denote $W$ as the vocabulary set or dictionary, where $w_{i}\in R^{|W|}$ is the one-hot word representation. Since both video and question content are sequential data with variant length, it is natural to choose the variant recurrent neural network called long-short term memory network (LSTM)~\cite{hochreiter1997long}.

Specifically, we learn the feature representation of both video and question by bidirectional LSTM, which consists of a forward LSTM and a backward LSTM~\cite{zhang2016learning}. The backward LSTM has the same network structure with the forward one while its input sequence is reversed. We denote the hidden state of the forward LSTM at time $t$ by ${\bf h}_{t}^{f}$, and the hidden state of the backward LSTM by ${\bf h}_{t}^{b}$. Thus, the hidden state of video ${\bf v}$ at time $t$ from bidirectional layer is denoted by ${\bf h}_{t}=[{\bf h}_{t}^{f},{\bf h}_{N-t+1}^{b}]$. The hidden states of video ${\bf v}$ is given by ${\bf h}_{v}=({\bf h}_{1},{\bf h}_{2},\ldots,{\bf h}_{N})$. We then denote the latent representation of question ${\bf q}$ from bidirectional layer by ${\bf h}_{q}$.

Using the notations above, the problem of video question answering is formulated as follows. Given the set of videos $V$, questions $Q$ and attributes $A$, our goal is to learn the attribute-augmented attention network such that when a certain question is issued, r-ANL can return the relevant answer for it based on the referenced video content. We present the details of the attribute-augmented attention network learning framework in Figure~\ref{fig:framework}.

\subsection{Attribute-Augmented Attention Network Learning}
In this section, we propose the attribute-augmented attention network to learn the joint representation of multimodal video content and detected attributes according to the question for both multiple choice and open-ended video question answering tasks.

We first employ a set of pre-trained attribute detectors to obtain the visual attributes for each frame in video ${\bf v}$, denoted as $a_{v,i}$~\cite{WangLW15, densecap, zhang2013attribute}. Each attribute ${\bf w}_{j}\in a_{v,i}$ corresponds to one entry in the vocabulary set $W$. We then obtain the representation for attribute set by $f(a_{v,i})=\frac{1}{|a_{v,i}|}\sum_{{\bf w}_{j}\in a_{v,i}}{\bf T}_{w}{\bf w}_{j}$, where ${\bf T}_{w}$ is the embedding matrix for attribute representation and $|a_{v,i}|$ is the size of attribute set for the $i$-th frame. We thus learn the joint representation of multimodal attributes and frame representation by $g_{{\bf v}_{i}}(a_{v,i})={\bf h}_{i}\otimes f(a_{v,i})$, where $\otimes$ is the element-wise product and ${\bf h}_{i}$ is from bidirectional layer at time $i$.

Inspired by the temporal attention mechanism, we introduce the attribute-augmented attention network to learn the attribute-augmented video representation according to the question for video question answering. Given the question ${\bf q}$ and the $i$-th frame of video ${\bf v}$, the temporal attention score $s_{qi}$ is given by:
\begin{eqnarray}
s_{qi} =  tanh({\bf W}_{q}{\bf h}_{q}+{\bf W}_{v}g_{{\bf v}_{i}}(a_{v,i})+{\bf b}_{t}),
\end{eqnarray}
where ${\bf W}_{q}$ and ${\bf W}_{v}$ are parameter matrices and ${\bf b}_{t}$ is bias vector. The ${\bf h}_{q}$ denotes the latent representation of question ${\bf q}$ and $g_{{\bf v}_{i}}(a_{v,i})$ is attribute-augmented latent representation of the $i$-th frame from bidirectional LSTM networks, respectively. For each frame ${\bf v}_{i}$, the activations in temporal dimension by the softmax function is given by $\alpha_{v,i}=\frac{\exp(s_{qi})}{\sum_{i=1}^{N}\exp(s_{qi})}$, which is the normalization of the temporal attention score. Thus, the temporally attended video representation according to question ${\bf q}$ is given by $m_{{\bf q}}({\bf v})=\sum_{i=1}^{N}\alpha_{v,i}g_{{\bf v}_{i}}(a_{v,i})$.

We then incorporate the multi-step reasoning process for the proposed attribute-augmented attention networks to further improve the performance of question-oriented video representation for video question answering. Given the attribute-augmented attention network $m_{{\bf q}}({\bf v})$, video ${\bf v}$ and question ${\bf q}$, the attribute-augmented attention network learning with multi-step reasoning process is given by:
\begin{eqnarray}
{\bf z}_{r} &=& {\bf z}_{r-1} + m_{{\bf z}_{r-1}}({\bf v}),\nonumber\\
{\bf z}_{0} &=& {\bf q},
\end{eqnarray}
which is recursively updated. The joint question-oriented video representation is then returned after the $R$-th reasoning process update, given by ${\bf z}_{R}$. The learning process of reasoning attribute-augmented attention networks in case of $r=2$ is illustrated in Figure~\ref{fig:framework}.

We next present the objective function of our method for both multiple-choice and open-ended video question answering tasks. For training the model for multiple-choice task, we model video question answering as a classification problem with pre-defined classes. Given the updated joint question-oriented video representation ${\bf z}_{R}$, a softmax function is then employed to classify ${\bf z}_{R}$ into one of the possible answers as
\begin{eqnarray}
p_{y} = softmax(W_{y}{\bf z}_{R}+b_{y}),
\end{eqnarray}
where ${\bf W}_{y}$ is the parameter matrix and ${\bf b}_{y}$ is the bias vector. On the other hand, for training the model for open-ended video question answering, we employ the LSTM decoder $d(\cdot)$ to generate free-form answers based on the updated joint question-oriented video representation ${\bf z}_{R}$. Given video ${\bf v}$, question ${\bf q}$ and ground-truth answer ${\bf y}=({\bf y}_{1},{\bf y}_{2},\ldots, {\bf y}_{M})$ and the generated answer ${\bf o}=({\bf o}_{1},{\bf o}_{2},\ldots,{\bf o}_{M})$, the loss function $\mathcal{L}(d({\bf z}_{R}),{\bf y})$ is given by:
\begin{eqnarray}
\mathcal{L}(d({\bf z}_{R}),{\bf y}) = \sum_{i=1}^{M}{\bf 1}[{\bf y}_{i}\neq {\bf o}_{i}],
\end{eqnarray}
where ${\bf 1}[\cdot]$ is the indicator function. We denote all the model coefficients including neural network parameters and the result embeddings by $\Theta$. Therefore, the objective function in our learning process is given by
\begin{eqnarray}
\min_{\Theta}\mathcal{L}(\Theta) = \mathcal{L}_{\Theta}+ \lambda\|\Theta\|^{2},
\end{eqnarray}
where $\lambda$ is the trade-off parameter between the training loss and regularization. To optimize the objective function, we employ the stochastic gradient descent (SGD) with the diagonal variant of AdaGrad.

\section{Experiments}

\subsection{Data Preparation}

\begin{table}
\small
\centering
\caption{Summary of Dataset}\label{table:dataset}
\begin{tabular}{|c|c|c|c|}
\hline
\multirow{2}{*}{Data Splitting} & \multicolumn{3}{c|}{Question Types}\\
\cline{2-4}
  &What&  Who&  Other\\
\hline
train&  57,385& 27,316& 3,649\\
valid &3,495& 2,804&  182\\
test& 2,489 &2,004& 97\\
\hline
\end{tabular}
\end{table}

We construct the dataset of video question-answering from the YouTube2Text data~\cite{guadarrama2013youtube2text} with natural language descriptions, which consists of 1,987 videos and 122,708 descriptions. Following the state-of-the-art question generation method, we generate the question-answer pairs from the video descriptions. Following the existing visual question answering approaches~\cite{antol2015vqa}, we generate three types of questions, which are related to the what, who and other queries for the video. We split the generated dataset into three parts: the training, the validation and the testing sets. The three types of video question-answering pairs used for the experiments are summarized in Table~\ref{table:dataset}. The dataset will be provided later.

We then preprocess the video question-answering dataset as follows. We first sample 40 frames from each video and then resize each frame to 300$\times$300. We extract the visual representation of each frame by the pretrained ResNet~\cite{he2016deep}, and take the 2,048-dimensional feature vector for each frame~\cite{ZhaoHCZNZ16}. We employ the pretrained word2vec model to extract the semantic representation of questions and answers~\cite{ZhaoZHN15}. Specifically, the size of vocabulary set is 6,500 and the dimension of word vector is set to 256. For training model for open-ended video question answering task, we add a token $<$eos$>$ to mark the end of the answer phrase, and take the token $<$Unk$>$ for the out-of-vocabulary word.

\subsection{Performance Comparisons}

\begin{table*}[t]
\footnotesize
\centering
\caption{Experimental results on both open-ended and multiple-choice video question answering tasks.}\label{table:exp}
\begin{tabular}{|c|c|c|c|c|c|c|c|c|}
\hline
\multirow{3}{*}{Method} & \multicolumn{4}{c|}{Open-ended VQA task question type}& \multicolumn{4}{c|}{Multiple-choice VQA task question type}\\
\cline{2-9}
  & What  & Who & Other & Total accuracy  & What  & Who & Other & Total accuracy  \\
\hline
VQA+&0.2097&0.2486&0.7010&0.386&0.5998&0.3071&0.8144&0.574\\
SAN+&0.168&0.224&0.722&0.371&0.582&0.288&0.804&0.558\\
r-ANL$_{(-a)}$&0.164&0.231&0.784&0.393&0.550&0.288&0.825&0.554\\
r-ANL$_{(1)}$&0.179&0.235&0.701&0.372&0.582&0.261&0.825&0.556\\
r-ANL$_{(2)}$&0.158&0.249&0.794&0.400&0.603&0.285&0.825&0.571\\
r-ANL$_{(3)}$&{\bf 0.216}&{\bf 0.294}&{\bf 0.804}&{\bf 0.438}&{\bf 0.633}&{\bf 0.364}&{\bf 0.845}&{\bf 0.614}\\
\hline
\end{tabular}
\end{table*}

We evaluate the performance of our proposed r-ANL method on both multiple-choice and open-ended video question answering tasks using the evaluation criteria of Accuracy. Given the testing question ${\bf q}\in Q_{t}$ and video ${\bf v}\in V_{t}$ with the groundtruth answer ${\bf y}$, we denote the predicted answer by our r-ANL method by ${\bf o}$. We then introduce the evaluation criteria of $Accuracy$ below:
\begin{eqnarray}
Accuracy = \frac{1}{|Q_{t}|}\sum_{{\bf q}\in Q_{t},{\bf v}\in V_{t}}(1-\prod_{i=1}^{K}{\bf 1}[{\bf y}_{i}\neq {\bf o}_{i}]),\nonumber
\end{eqnarray}
where $Accuracy=1$ (best) means that the generated answer and the ground-truth ones are exactly the same, while $Accuracy = 0$ means the opposite. When we performance the multiple-choice video question answering task, we set the value of $K$ to 1.

We extend the existing image question answering methods as the baseline algorithms for the problem of video question answering.
\begin{itemize}
\item {\bf VQA+} method is the extension of VQA algorithm~\cite{antol2015vqa}, where we add the mean-pooling layer that obtains the joint video representation from ResNet-based frame features, and then computes the joint representation of question embedding and video representation by their element-wise multiplication for generating open-ended answers.
 \item {\bf SAN+} method is the incremental algorithm based on stacked attention networks~\cite{yang2016stacked}, where we add the LSTM network to fuse the sequential representation of video frames for video question answering.
\end{itemize}
Unlike the previous visual question answering works, our r-ANL method learns the question-oriented video question with multiple reasoning process for the problem of video question answering. To study the effectiveness of attribute-augmented mechanism in our attention network, we evaluate our method with the one without attributes, denoted as r-ANL$_{(-a)}$. To exploit the effect of reasoning process, we denote our r-ANL method with $r$ reasoning steps by r-ANL$_{(r)}$. The input words of our method are initialized by pre-trained word embeddings with size of 256, and weights of LSTMs are randomly by a Gaussian distribution with zero mean.

\begin{figure}
\centering
\includegraphics[width=0.35\textwidth,height=0.15\textwidth]{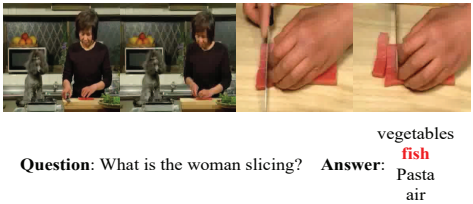}
\caption{Experimental results of both open-ended and multiple-choice video question answering}\label{fig:demo exp}
\end{figure}

Table~\ref{table:exp} shows the overall experimental results of the methods on both open-ended and multiple-choice video question answering tasks with different types of questions.
The hyperparameters and parameters which achieve the best performance on the validation set are chosen to conduct the testing evaluation. We report the average value of all the methods on three evaluation criteria. We give an example of the experimental results by our method in Figure~\ref{fig:demo exp}.

\section{Conclusion}

In this paper, we study the problem of video question answering from the viewpoint of attribute-augmented attention network learning. We first propose the attribute-augmented method that learns the joint representation of visual frame and textual attributes. We then develop the attribute-augmented attention network to learn the question-oriented video representation for question answering. We next incorporate the multi-step reasoning process to our proposed attention network that further improve the performance of the method for the problem. We construct a large-scale video question answering dataset and evaluate the effectiveness of our proposed method through extensive experiments.

\begin{acks}

  The work is supported by the National Natural
    Science Foundation of China 
    under Grant No.61572431 and No.61602405.
    It is also supported by the Fundamental Research Funds for the Central Universities 2016QNA5015,
    Zhejiang Natural Science Foundation under Grant LZ17F020001,  
    and the China Knowledge Centre for Engineering Sciences and Technology.
\end{acks}

\bibliographystyle{ACM-Reference-Format}
\bibliography{ref} 


\begin{thebibliography}{00}


\ifx \showCODEN    \undefined \def \showCODEN     #1{\unskip}     \fi
\ifx \showDOI      \undefined \def \showDOI       #1{#1}\fi
\ifx \showISBNx    \undefined \def \showISBNx     #1{\unskip}     \fi
\ifx \showISBNxiii \undefined \def \showISBNxiii  #1{\unskip}     \fi
\ifx \showISSN     \undefined \def \showISSN      #1{\unskip}     \fi
\ifx \showLCCN     \undefined \def \showLCCN      #1{\unskip}     \fi
\ifx \shownote     \undefined \def \shownote      #1{#1}          \fi
\ifx \showarticletitle \undefined \def \showarticletitle #1{#1}   \fi
\ifx \showURL      \undefined \def \showURL       {\relax}        \fi
\providecommand\bibfield[2]{#2}
\providecommand\bibinfo[2]{#2}
\providecommand\natexlab[1]{#1}
\providecommand\showeprint[2][]{arXiv:#2}

\bibitem[\protect\citeauthoryear{Antol, Agrawal, Lu, Mitchell, Batra,
  Lawrence~Zitnick, and Parikh}{Antol et~al\mbox{.}}{2015}]%
        {antol2015vqa}
\bibfield{author}{\bibinfo{person}{Stanislaw Antol}, \bibinfo{person}{Aishwarya
  Agrawal}, \bibinfo{person}{Jiasen Lu}, \bibinfo{person}{Margaret Mitchell},
  \bibinfo{person}{Dhruv Batra}, \bibinfo{person}{C Lawrence~Zitnick}, {and}
  \bibinfo{person}{Devi Parikh}.} \bibinfo{year}{2015}\natexlab{}.
\newblock \showarticletitle{Vqa: Visual question answering}. In
  \bibinfo{booktitle}{{\em ICCV}}. \bibinfo{pages}{2425--2433}.
\newblock


\bibitem[\protect\citeauthoryear{Guadarrama, Krishnamoorthy, Malkarnenkar,
  Venugopalan, Mooney, Darrell, and Saenko}{Guadarrama et~al\mbox{.}}{2013}]%
        {guadarrama2013youtube2text}
\bibfield{author}{\bibinfo{person}{Sergio Guadarrama}, \bibinfo{person}{Niveda
  Krishnamoorthy}, \bibinfo{person}{Girish Malkarnenkar},
  \bibinfo{person}{Subhashini Venugopalan}, \bibinfo{person}{Raymond Mooney},
  \bibinfo{person}{Trevor Darrell}, {and} \bibinfo{person}{Kate Saenko}.}
  \bibinfo{year}{2013}\natexlab{}.
\newblock \showarticletitle{Youtube2text: Recognizing and describing arbitrary
  activities using semantic hierarchies and zero-shot recognition}. In
  \bibinfo{booktitle}{{\em ICCV}}. \bibinfo{pages}{2712--2719}.
\newblock


\bibitem[\protect\citeauthoryear{Gupta and Jain}{Gupta and Jain}{1997}]%
        {gupta1997visual}
\bibfield{author}{\bibinfo{person}{Amarnath Gupta} {and}
  \bibinfo{person}{Ramesh Jain}.} \bibinfo{year}{1997}\natexlab{}.
\newblock \showarticletitle{Visual information retrieval}.
\newblock \bibinfo{journal}{{\it Commun. ACM}} \bibinfo{volume}{40},
  \bibinfo{number}{5} (\bibinfo{year}{1997}), \bibinfo{pages}{70--79}.
\newblock


\bibitem[\protect\citeauthoryear{He, Zhang, Ren, and Sun}{He
  et~al\mbox{.}}{2016b}]%
        {he2016deep}
\bibfield{author}{\bibinfo{person}{Kaiming He}, \bibinfo{person}{Xiangyu
  Zhang}, \bibinfo{person}{Shaoqing Ren}, {and} \bibinfo{person}{Jian Sun}.}
  \bibinfo{year}{2016}\natexlab{b}.
\newblock \showarticletitle{Deep residual learning for image recognition}. In
  \bibinfo{booktitle}{{\em CVPR}}. \bibinfo{pages}{770--778}.
\newblock


\bibitem[\protect\citeauthoryear{He, Gao, Kan, and Wang}{He
  et~al\mbox{.}}{2017a}]%
        {HeGKW17}
\bibfield{author}{\bibinfo{person}{Xiangnan He}, \bibinfo{person}{Ming Gao},
  \bibinfo{person}{Min{-}Yen Kan}, {and} \bibinfo{person}{Dingxian Wang}.}
  \bibinfo{year}{2017}\natexlab{a}.
\newblock \showarticletitle{BiRank: Towards Ranking on Bipartite Graphs}.
\newblock \bibinfo{journal}{{\em {IEEE} Trans. Knowl. Data Eng.\/}}
  (\bibinfo{year}{2017}), \bibinfo{pages}{57--71}.
\newblock


\bibitem[\protect\citeauthoryear{He, Liao, Zhang, Nie, Hu, and Chua}{He
  et~al\mbox{.}}{2017b}]%
        {HeLZNHC17}
\bibfield{author}{\bibinfo{person}{Xiangnan He}, \bibinfo{person}{Lizi Liao},
  \bibinfo{person}{Hanwang Zhang}, \bibinfo{person}{Liqiang Nie},
  \bibinfo{person}{Xia Hu}, {and} \bibinfo{person}{Tat{-}Seng Chua}.}
  \bibinfo{year}{2017}\natexlab{b}.
\newblock \showarticletitle{Neural Collaborative Filtering}. In
  \bibinfo{booktitle}{{\em WWW}}. \bibinfo{pages}{173--182}.
\newblock


\bibitem[\protect\citeauthoryear{He, Zhang, Kan, and Chua}{He
  et~al\mbox{.}}{2016a}]%
        {HeZKC16}
\bibfield{author}{\bibinfo{person}{Xiangnan He}, \bibinfo{person}{Hanwang
  Zhang}, \bibinfo{person}{Min{-}Yen Kan}, {and} \bibinfo{person}{Tat{-}Seng
  Chua}.} \bibinfo{year}{2016}\natexlab{a}.
\newblock \showarticletitle{Fast Matrix Factorization for Online Recommendation
  with Implicit Feedback}. In \bibinfo{booktitle}{{\em SIGIR}}.
  \bibinfo{pages}{549--558}.
\newblock


\bibitem[\protect\citeauthoryear{Hochreiter and Schmidhuber}{Hochreiter and
  Schmidhuber}{1997}]%
        {hochreiter1997long}
\bibfield{author}{\bibinfo{person}{Sepp Hochreiter} {and}
  \bibinfo{person}{J{\"u}rgen Schmidhuber}.} \bibinfo{year}{1997}\natexlab{}.
\newblock \showarticletitle{Long short-term memory}.
\newblock \bibinfo{journal}{{\em Neural computation\/}} \bibinfo{volume}{9},
  \bibinfo{number}{8} (\bibinfo{year}{1997}), \bibinfo{pages}{1735--1780}.
\newblock


\bibitem[\protect\citeauthoryear{Johnson, Karpathy, and Fei-Fei}{Johnson
  et~al\mbox{.}}{2016}]%
        {densecap}
\bibfield{author}{\bibinfo{person}{Justin Johnson}, \bibinfo{person}{Andrej
  Karpathy}, {and} \bibinfo{person}{Li Fei-Fei}.}
  \bibinfo{year}{2016}\natexlab{}.
\newblock \showarticletitle{DenseCap: Fully Convolutional Localization Networks
  for Dense Captioning}. In \bibinfo{booktitle}{{\em CVPR}}.
\newblock


\bibitem[\protect\citeauthoryear{Luo, Ni, Yan, and Wang}{Luo
  et~al\mbox{.}}{2016}]%
        {LuoNYW16}
\bibfield{author}{\bibinfo{person}{Changzhi Luo}, \bibinfo{person}{Bingbing
  Ni}, \bibinfo{person}{Shuicheng Yan}, {and} \bibinfo{person}{Meng Wang}.}
  \bibinfo{year}{2016}\natexlab{}.
\newblock \showarticletitle{Image Classification by Selective Regularized
  Subspace Learning}.
\newblock \bibinfo{journal}{{\em {IEEE} Trans. Multimedia\/}}
  (\bibinfo{year}{2016}), \bibinfo{pages}{40--50}.
\newblock


\bibitem[\protect\citeauthoryear{Nie, Wang, Gao, Zha, and Chua}{Nie
  et~al\mbox{.}}{2013}]%
        {NieWGZC13}
\bibfield{author}{\bibinfo{person}{Liqiang Nie}, \bibinfo{person}{Meng Wang},
  \bibinfo{person}{Yue Gao}, \bibinfo{person}{Zheng{-}Jun Zha}, {and}
  \bibinfo{person}{Tat{-}Seng Chua}.} \bibinfo{year}{2013}\natexlab{}.
\newblock \showarticletitle{Beyond Text {QA:} Multimedia Answer Generation by
  Harvesting Web Information}.
\newblock \bibinfo{journal}{{\em {IEEE} Trans. Multimedia\/}}
  (\bibinfo{year}{2013}), \bibinfo{pages}{426--441}.
\newblock


\bibitem[\protect\citeauthoryear{Nie, Wang, Zha, Li, and Chua}{Nie
  et~al\mbox{.}}{2011}]%
        {NieWZLC11}
\bibfield{author}{\bibinfo{person}{Liqiang Nie}, \bibinfo{person}{Meng Wang},
  \bibinfo{person}{Zheng{-}Jun Zha}, \bibinfo{person}{Guangda Li}, {and}
  \bibinfo{person}{Tat{-}Seng Chua}.} \bibinfo{year}{2011}\natexlab{}.
\newblock \showarticletitle{Multimedia answering: enriching text {QA} with
  media information}. In \bibinfo{booktitle}{{\em SIGIR}}.
  \bibinfo{pages}{695--704}.
\newblock


\bibitem[\protect\citeauthoryear{Nie, Yan, Wang, Hong, and Chua}{Nie
  et~al\mbox{.}}{2012}]%
        {NieYWHC12}
\bibfield{author}{\bibinfo{person}{Liqiang Nie}, \bibinfo{person}{Shuicheng
  Yan}, \bibinfo{person}{Meng Wang}, \bibinfo{person}{Richang Hong}, {and}
  \bibinfo{person}{Tat{-}Seng Chua}.} \bibinfo{year}{2012}\natexlab{}.
\newblock \showarticletitle{Harvesting visual concepts for image search with
  complex queries}. In \bibinfo{booktitle}{{\em ACM MM}}.
  \bibinfo{pages}{59--68}.
\newblock


\bibitem[\protect\citeauthoryear{Wang, Hong, Li, Zha, Yan, and Chua}{Wang
  et~al\mbox{.}}{2012a}]%
        {WangHLZYC12}
\bibfield{author}{\bibinfo{person}{Meng Wang}, \bibinfo{person}{Richang Hong},
  \bibinfo{person}{Guangda Li}, \bibinfo{person}{Zheng{-}Jun Zha},
  \bibinfo{person}{Shuicheng Yan}, {and} \bibinfo{person}{Tat{-}Seng Chua}.}
  \bibinfo{year}{2012}\natexlab{a}.
\newblock \showarticletitle{Event Driven Web Video Summarization by Tag
  Localization and Key-Shot Identification}.
\newblock \bibinfo{journal}{{\em {IEEE} Trans. Multimedia\/}}
  (\bibinfo{year}{2012}), \bibinfo{pages}{975--985}.
\newblock


\bibitem[\protect\citeauthoryear{Wang, Li, Tao, Lu, and Wu}{Wang
  et~al\mbox{.}}{2012b}]%
        {WangLTLW12}
\bibfield{author}{\bibinfo{person}{Meng Wang}, \bibinfo{person}{Hao Li},
  \bibinfo{person}{Dacheng Tao}, \bibinfo{person}{Ke Lu}, {and}
  \bibinfo{person}{Xindong Wu}.} \bibinfo{year}{2012}\natexlab{b}.
\newblock \showarticletitle{Multimodal Graph-Based Reranking for Web Image
  Search}.
\newblock \bibinfo{journal}{{\em {IEEE} Trans. Image Processing\/}}
  (\bibinfo{year}{2012}), \bibinfo{pages}{4649--4661}.
\newblock


\bibitem[\protect\citeauthoryear{Wang, Liu, and Wu}{Wang et~al\mbox{.}}{2015}]%
        {WangLW15}
\bibfield{author}{\bibinfo{person}{Meng Wang}, \bibinfo{person}{Xueliang Liu},
  {and} \bibinfo{person}{Xindong Wu}.} \bibinfo{year}{2015}\natexlab{}.
\newblock \showarticletitle{Visual Classification by -Hypergraph Modeling}.
\newblock \bibinfo{journal}{{\em {IEEE} Trans. Knowl. Data Eng.\/}}
  (\bibinfo{year}{2015}), \bibinfo{pages}{2564--2574}.
\newblock


\bibitem[\protect\citeauthoryear{Yang, He, Gao, Deng, and Smola}{Yang
  et~al\mbox{.}}{2016}]%
        {yang2016stacked}
\bibfield{author}{\bibinfo{person}{Zichao Yang}, \bibinfo{person}{Xiaodong He},
  \bibinfo{person}{Jianfeng Gao}, \bibinfo{person}{Li Deng}, {and}
  \bibinfo{person}{Alex Smola}.} \bibinfo{year}{2016}\natexlab{}.
\newblock \showarticletitle{Stacked attention networks for image question
  answering}. In \bibinfo{booktitle}{{\em CVPR}}. \bibinfo{pages}{21--29}.
\newblock


\bibitem[\protect\citeauthoryear{Zhang, Shang, Luan, Wang, and Chua}{Zhang
  et~al\mbox{.}}{2016}]%
        {zhang2016learning}
\bibfield{author}{\bibinfo{person}{Hanwang Zhang}, \bibinfo{person}{Xindi
  Shang}, \bibinfo{person}{Huanbo Luan}, \bibinfo{person}{Meng Wang}, {and}
  \bibinfo{person}{Tat-Seng Chua}.} \bibinfo{year}{2016}\natexlab{}.
\newblock \showarticletitle{Learning from collective intelligence: Feature
  learning using social images and tags}.
\newblock \bibinfo{journal}{{\em {IEEE} Trans. Multimedia\/}}
  \bibinfo{volume}{13} (\bibinfo{year}{2016}).
\newblock


\bibitem[\protect\citeauthoryear{Zhang, Zha, Yang, Yan, Gao, and Chua}{Zhang
  et~al\mbox{.}}{2013}]%
        {zhang2013attribute}
\bibfield{author}{\bibinfo{person}{Hanwang Zhang}, \bibinfo{person}{Zheng-Jun
  Zha}, \bibinfo{person}{Yang Yang}, \bibinfo{person}{Shuicheng Yan},
  \bibinfo{person}{Yue Gao}, {and} \bibinfo{person}{Tat-Seng Chua}.}
  \bibinfo{year}{2013}\natexlab{}.
\newblock \showarticletitle{Attribute-augmented semantic hierarchy: towards
  bridging semantic gap and intention gap in image retrieval}. In
  \bibinfo{booktitle}{{\em ACM MM}}. ACM, \bibinfo{pages}{33--42}.
\newblock


\bibitem[\protect\citeauthoryear{Zhao, He, Cai, Zhang, Ng, and Zhuang}{Zhao
  et~al\mbox{.}}{2016}]%
        {ZhaoHCZNZ16}
\bibfield{author}{\bibinfo{person}{Zhou Zhao}, \bibinfo{person}{Xiaofei He},
  \bibinfo{person}{Deng Cai}, \bibinfo{person}{Lijun Zhang},
  \bibinfo{person}{Wilfred Ng}, {and} \bibinfo{person}{Yueting Zhuang}.}
  \bibinfo{year}{2016}\natexlab{}.
\newblock \showarticletitle{Graph Regularized Feature Selection with Data
  Reconstruction}.
\newblock \bibinfo{journal}{{\em {IEEE} Trans. Knowl. Data Eng.\/}}
  (\bibinfo{year}{2016}), \bibinfo{pages}{689--700}.
\newblock


\bibitem[\protect\citeauthoryear{Zhao, Lu, Zheng, Cai, He, and Zhuang}{Zhao
  et~al\mbox{.}}{2017}]%
        {ZhaoLZCHZ17}
\bibfield{author}{\bibinfo{person}{Zhou Zhao}, \bibinfo{person}{Hanqing Lu},
  \bibinfo{person}{Vincent~W. Zheng}, \bibinfo{person}{Deng Cai},
  \bibinfo{person}{Xiaofei He}, {and} \bibinfo{person}{Yueting Zhuang}.}
  \bibinfo{year}{2017}\natexlab{}.
\newblock \showarticletitle{Community-Based Question Answering via Asymmetric
  Multi-Faceted Ranking Network Learning}. In \bibinfo{booktitle}{{\em AAAI}}.
  \bibinfo{pages}{3532--3539}.
\newblock


\bibitem[\protect\citeauthoryear{Zhao, Yang, Cai, He, and Zhuang}{Zhao
  et~al\mbox{.}}{2016}]%
        {ZhaoYCHZ16}
\bibfield{author}{\bibinfo{person}{Zhou Zhao}, \bibinfo{person}{Qifan Yang},
  \bibinfo{person}{Deng Cai}, \bibinfo{person}{Xiaofei He}, {and}
  \bibinfo{person}{Yueting Zhuang}.} \bibinfo{year}{2016}\natexlab{}.
\newblock \showarticletitle{Expert Finding for Community-Based Question
  Answering via Ranking Metric Network Learning}. In \bibinfo{booktitle}{{\em
  IJCAI}}. \bibinfo{pages}{3000--3006}.
\newblock


\bibitem[\protect\citeauthoryear{Zhao, Zhang, He, and Ng}{Zhao
  et~al\mbox{.}}{2015}]%
        {ZhaoZHN15}
\bibfield{author}{\bibinfo{person}{Zhou Zhao}, \bibinfo{person}{Lijun Zhang},
  \bibinfo{person}{Xiaofei He}, {and} \bibinfo{person}{Wilfred Ng}.}
  \bibinfo{year}{2015}\natexlab{}.
\newblock \showarticletitle{Expert Finding for Question Answering via Graph
  Regularized Matrix Completion}.
\newblock \bibinfo{journal}{{\em {IEEE} Trans. Knowl. Data Eng.\/}}
  (\bibinfo{year}{2015}), \bibinfo{pages}{993--1004}.
\newblock


\end{thebibliography}

\end{document}